\documentclass[runningheads]{llncs2}
\usepackage{graphicx}
\usepackage{amsmath} 
\usepackage{amssymb}  
\usepackage{hyperref}
\usepackage[table]{xcolor}
\usepackage{float}
\usepackage{setspace}
\usepackage{fancyhdr}
\usepackage{multirow}
\usepackage{caption}

\DeclareMathOperator{\EX}{\mathbb{E}}
\begin{document}

\title{Variance Loss \\ in Variational Autoencoders
}


\author{Andrea Asperti}

\institute{
University  of Bologna\\
Department of Informatics: Science and Engineering (DISI)\\
              \email{andrea.asperti@unibo.it}           
}


\maketitle

\begin{abstract}
In this article, we highlight what appears to be major issue of Variational
Autoencoders (VAEs), evinced from an extensive experimentation with different networks 
architectures and datasets: the variance of generated 
data is significantly lower than that of
training data. Since generative models are usually evaluated with metrics 
such as the Fr\'echet Inception Distance (FID) that compare
the distributions of (features of) real versus generated images, 
the variance loss typically results in degraded scores.
This problem is particularly relevant in a two stage setting \cite{TwoStage}, where a second VAE is used to sample in the latent space of the first VAE. The minor variance
creates a mismatch between the actual distribution of latent variables and those
generated by the second VAE, that hinders the beneficial effects of the second stage.
Renormalizing the output of the second VAE towards the expected normal spherical
distribution, we obtain a sudden burst in the quality of generated samples, as also testified in terms of FID.
\end{abstract}

\section{Introduction}
\label{sec:intro}
Since their introduction (\cite{Kingma13,RezendeMW14}), Variational Autoencoders 
(VAEs) have rapidly become one of the most popular frameworks for generative modeling.
Their appeal mostly derives from the strong probabilistic foundation; moreover, 
they are traditionally reputed for granting more stable training than 
Generative Adversarial Networks (GANs) (\cite{GAN}).


However, the behaviour of Variational Autoencoders is still far from satisfactory, and there
are a lot of well known theoretical and practical challenges that still hinder 
this generative paradigm. We may roughly identify four main (interrelated) topics that 
have been 
addressed so far:
\begin{description}
\item[balancing issue] 
\cite{Bowman15,autoregressive16,beta-vae17,understanding-beta-vae18,TwoStage,balancing} a major problem of VAE is the difficulty to find a good compromise between 
sampling quality and reconstruction quality. The VAE loss function is a combination
of two terms with somehow contrasting effects: the log-likelihood, aimed to reduce
the reconstruction error, and the Kullback-Leibler divergence, acting as a 
regularizer of the latent space with the final purpose to improve generative sampling (see Section~\ref{sec:vae} for details).
Finding a good balance between these components during training is a complex and delicate issue;
\item[variable collapse phenomenon]  \cite{BurdaGS15,overpruning17,sparsity,Trippe18,TwoStage}. The KL-divergence component of the VAE loss function typically induces a parsimonious use of latent variables, some of which
may be altogether neglected by the decoder, possibly resulting in a under-exploitation
of the network capacity; if this is a beneficial side effect or regularization (sparsity),
or an issue to be solved (overpruning), it is still debated;
\item[training issues] VAE approximate expectations through sampling during training that could cause an increased variance in gradients 
(\cite{BurdaGS15,TuckerMMLS17}); this and other issues require some attention in the initialization, validation, and annealing
of hyperparameters (\cite{Bowman15,beta-vae17,resampledPriors})
\item[aggregate posterior vs. expected prior mismatch] 
\cite{autoregressive16,TwoStage,aboutVAE,deterministic} even after a satisfactory 
convergence of training, there is no guarantee that the learned aggregated posterior
distribution will match the latent prior. This may be due to the choice of an overly simplistic prior distribution; alternatively, the issue can e.g. be addressed by learning the actual distribution, either via a second VAE or by ex-post estimation by means of different techniques.
\end{description}

The main contribution of this article is to highlight an additional issue that,
at the best of our knowledge, has never been pointed out so far: 
the variance of generated data is significantly lower than that of training data. 

This resulted from a long series of experiments we did with a large variety of
different architectures and datasets. The variance loss is systematic, although
its extent may vary, and looks roughly proportional to the reconstruction loss.

The problem is relevant because generative models are traditionally evaluated with metrics 
such as the popular Fr\'echet Inception Distance (FID)  that compare
the distributions of (features of) real versus generated images: any bias 
in generated data usually results in a severe penalty in terms of FID score.

The variance loss is particularly serious in a two stage setting \cite{TwoStage}, where we
use a second VAE to sample in the latent space of the first VAE. The reduced variance
induces a mismatch between the actual distribution of latent variables and those
generated by the second VAE, substantially hindering the beneficial effects of the second stage.

We address the issue by a simple renormalization of the
generated data to match the expected variance (that should be 1, in case of a two stage VAE). This simple expedient, in combination with a
new balancing technique for the VAE loss function discussed in a different article \cite{balancing}, are the basic ingredients that 
permitted us to get the {\em best FID scores} ever achieved with variational techniques over traditional datasets such as CIFAR-10 and CelebA.

The cause of the reduced variance is not easy to identify. A plausible explanation 
is the following. It is well known that, in presence of multimodal output, the mean
square error objective typically results in blurriness, due to averaging 
(see \cite{GANTutorial}). 

Variational Autoencoders are intrinsically multimodal, due to the sampling process during
training, comporting averaging around the input data $X$ in the data manifold, and finally resulting in 
the blurriness so typical of Variational Autoencoders \cite{DosovitskiyB16}. The reduced variance is
just a different facet of the same phenomenon: averaging on the data manifold eventually
reduces the variance of data, due to Jensen's inequality.

The structure of the article is the following. Section~\ref{sec:vae} contains a short introduction 
to Variational Autoencoders from an operational perspective, focusing on the regularization effect of the Kullback-Leibler component of the loss function. In Section~\ref{sec:variance_loss}, we discuss the variance loss issue, relating it to a similar
problem of Principal Component Analysis, and providing experimental evidence of the phenomenon.
Section~\ref{sec:renormalization} is devoted to our approach to the variance loss, with
experimental results on CIFAR-10 and CelebA, two of the most common datasets in the field of generative modeling.
A summary of the content of the article and concluding remarks are given in Section~\ref{Sec:conclusions}.

\section{Variational Autoencoders}\label{sec:vae}
A Variational Autoencoder is composed by an encoder computing an 
{\em inference} distribution $Q(z|X)$, and a decoder, computing the posterior
probability $P(X|z)$. Supposing that $Q(z|X)$ has a Gaussian distribution
$N(\mu_z(X),\sigma_z(X))$ (different for each data $X$), computing it amounts
to compute its two first moments: so we expect the encoder to return the 
standard deviation $\sigma_z(X)$ in addition to the mean value $\mu_z(X)$. 

\hspace*{-.5cm}\begin{minipage}{6cm}
During decoding, instead of starting the reconstruction from $\mu_z(X)$, 
we sample around this point with the computed standard deviation:
\[\hat{z} = \mu_z(X) + \sigma_z(X)*\delta \]
where $\delta$ is a random normal noise (see Figure~\ref{fig:vae}). 
This may be naively understood as
a way to inject noise in the latent representation, with the aim to 
improve the robustness of the autoencoder; in fact, it has a much stronger
theoretical foundation, well addressed in the literature (see e.g. \cite{tutorial-VAE}).
Observe that sampling is outside the backpropagation flow; 
backpropagating the reconstruction error (typically, mean squared error), we correct the
current estimation of $\sigma_z(X)$, along with the estimation of $\mu(X)$.
\end{minipage}
\begin{minipage}{6cm}
\begin{center}
\includegraphics[width=.9\textwidth]{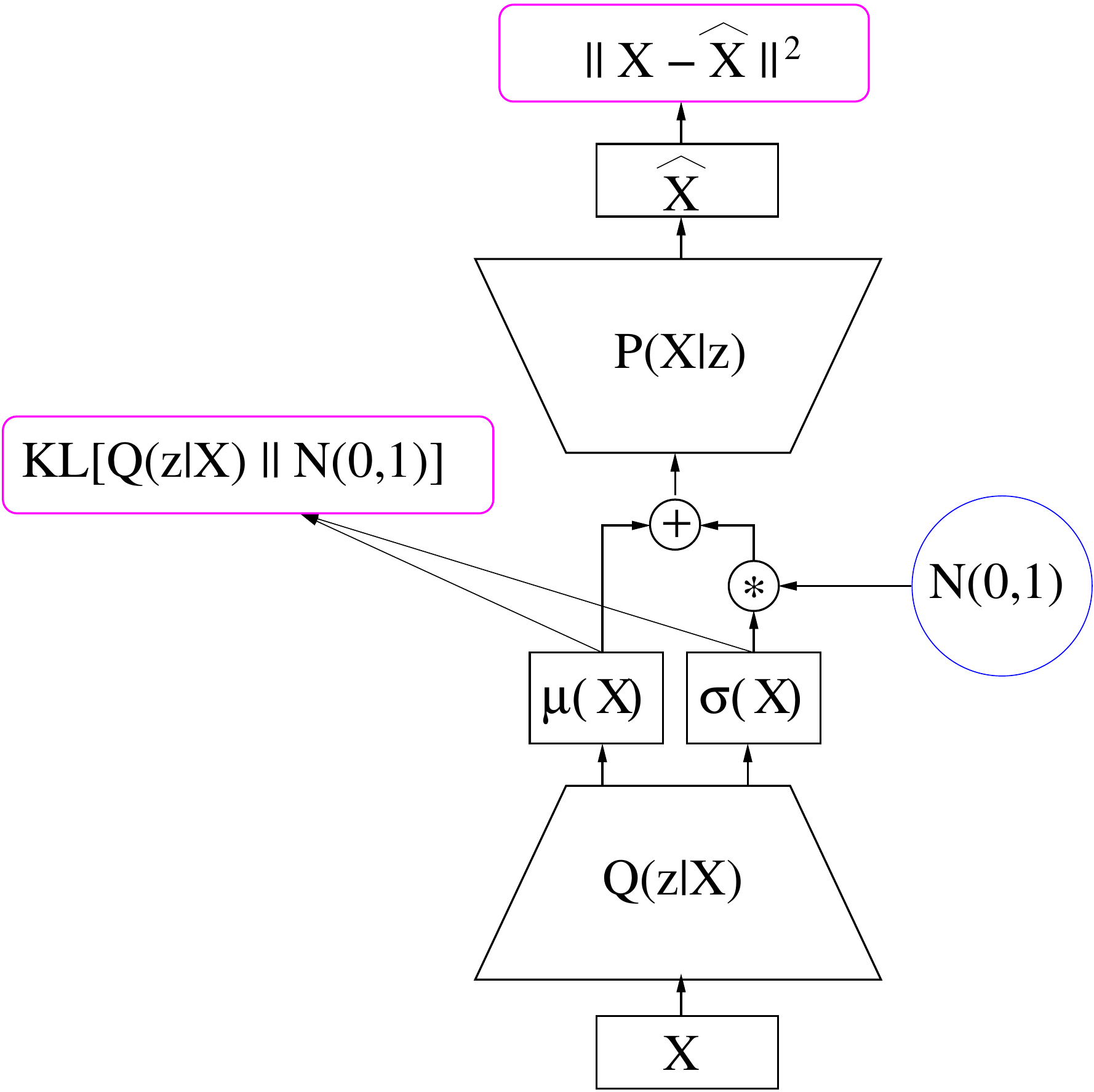}
\captionof{figure}{VAE architecture}
\label{fig:vae}
\end{center}
\end{minipage}

Without further constraints, $\sigma_z(X)$ would naturally collapse to 0:
as a matter of fact, $\mu_z(X)$ is the expected encoding, and the autoencoder
would have no reason to sample away from this value. The variational autoencoder adds an additional component to the loss function, preventing
$Q(z|X)$ from collapsing to a dirac distribution: specifically, we try 
to bring each $Q(z|X)$ close to the prior $P(z)$ distribution by minimizing
their Kullback-Leibler divergence $KL(Q(z|X)||P(z))$. 

If we average this quantity on all input data, and expand 
KL-divergence in terms of entropy, we get:
\begin{equation}\label{eq:averaging}
  \hspace{.5cm}\begin{array}{ll}
    \EX_{X}KL(Q(z|X)||P(z))\smallskip\\\smallskip
    = - \EX_{X} \mathcal{H}(Q(z|X)) + \EX_{X} \mathcal{H}(Q(z|X),P(z)) \\\smallskip
    = - \EX_{X} \mathcal{H}(Q(z|X)) + \EX_{X} \EX_{z \sim Q(z|X)}log P(z)  \\\smallskip
    = - \EX_{X} \mathcal{H}(Q(z|X)) + \EX_{z \sim Q(z)}log P(z)\\\smallskip
    = - \underbrace{\EX_{X} \mathcal{H}(Q(z|X))}_{\substack{\mbox{Avg. Entropy}\\\mbox{of } Q(z|X)}} +
          \underbrace{\mathcal{H}(Q(z),P(z))}_{\substack{\mbox{Cross-entropy }\\\mbox{of }Q(X) \mbox{ vs }P(z)}}
  \end{array}
  \end{equation}
  By minimizing the cross-entropy between the distributions we are pushing $Q(z)$ towards $P(z)$. Simultaneously, we aim to augment the entropy of each $Q(z|X)$; assuming $Q(z|X)$ is Gaussian, this amounts to 
  enlarge the variance, with the effect of improving the coverage of the latent space, essential for a good generative sampling. The price we
  have to pay is more overlapping, and hence more confusion, between the 
  encoding of different datapoints, likely resulting in a worse 
  reconstruction quality.
  
  \subsection{KL divergence in closed form}

We already supposed that $Q(X|z)$ has a Gaussian distribution 
$N(\mu_z(X),\sigma_z(X))$. Moreover, provided the decoder is sufficiently expressive, the shape of the prior distribution $P(z)$ can be arbitrary, and for simplicity it is usually assumed to be a normal distribution $P(z) = N(0,1)$.
The term $KL(Q(z|X)||P(z)$ is hence the KL-divergence between two Gaussian distributions $N(\mu_z(X),\sigma_z(X))$ and $N(1,0)$ which can be 
computed in closed form:
\begin{equation}\label{eq:closed_form}
\hspace{.3cm}\begin{array}{l}
    KL(N(\mu_z(X),\sigma_z(X)),N(0,1)) = \\
    \hspace{1cm}\frac{1}{2}(\mu_z(X)^2 + \sigma^2_z(X)-log(\sigma^2_z(X)) -1)
\end{array}
\end{equation}
The closed form helps to get some intuition on the way the regularizing effect of the KL-divergence is supposed to work.
The quadratic penalty $\mu_z(X)^2$ is centering the latent 
space around the origin; moreover, under the assumption to
fix the ratio between $\mu_z(X)$ and $\sigma_z(X)$ (rescaling
is an easy operation for a neural network) it is easy to prove
\cite{aboutVAE}
that expression~\ref{eq:closed_form} has a minimum when 
$\mu_z(X)^2 + \sigma_z(X)^2 =1$. So, we expect
\begin{equation}\label{eq:GMMmean}
\EX_X \mu(X) = 0
\end{equation}
and also, assuming \ref{eq:GMMmean}, and some further approximation (see \cite{aboutVAE} for details), 
 \begin{equation}\label{eq:var-law}
\EX_X \mu_z(X)^2 +  \EX_X \sigma_z^2(X)  = 1
\end{equation}
If we look at $Q(z) = \EX_X Q(z|X)$ as a Gaussian Mixture Model (GMM) composed by a different Gaussian $Q(z|X)$ for each $X$,
the two previous equations express the two moments of the GMM,
confirming that they coincide with those of a normal prior.
Equation~\ref{eq:var-law}, that we call {\em variance law}, provides a simple sanity check to ensure
that the regularization effect of the KL-divergence is working
as expected. 

Of course, even if two first moments of the aggregated inference distribution $Q(z)$ are 0 and 1, it could still be very far from 
a Normal distribution. The possible mismatching between $Q(z)$ and the expected prior $P(z)$ is likely the most problematic aspect
of VAEs since, as observed by several authors \cite{ELBOsurgery,rosca2018distribution,aboutVAE}, it could
compromise the whole generative framework. Possible
approaches consist in revising the VAE objective by 
encouraging the aggregated inference distribution to match $P(z)$
\cite{WAE} or by exploiting more complex priors \cite{autoregressive,Vamp,resampledPriors}.

An interesting alternative addressed in \cite{TwoStage} is that
of training a second VAE to learn an accurate approximation of 
$Q(z)$; samples from a Normal distribution are first used to generate samples of $Q(z)$, that are then fed to the actual generator of data points. 
Similarly, in \cite{deterministic}, the authors try to give an ex-post estimation of $Q(z)$, e.g. imposing 
a distribution with a sufficient complexity (they consider a combination of 10 Gaussians, reflecting the ten categories of MNIST and Cifar10). 

These two works provide the current state of the art in generative
frameworks based on variational techniques (hence, excluding models based on adversarial training), so we shall mostly compare with them. 

\section{The variance loss issue}\label{sec:variance_loss}
Autoencoders, and especially variational ones, seems to suffer from
a systematic loss of variance of reconstructed/generated data with respect to source
data. Suppose to have a training set {\bf $X$} of $n$ data, each one with $m$ features, and let {\bf $\hat{X}$} be the
corresponding set of reconstructed data. We measure the (mean) variance loss as the mean over data (that is over the the first axis) of the
differences of the variances of the features (i.e. over the second, default, axis):
\[\mbox{mean(var({\bf $X$}) - var({\bf $\hat X$}))}\]
Not only this quantity is always positive, but it is also approximately equal to the mean squared error (mse) between {\bf $X$} and {\bf $\hat X$}:
\[\mbox{mse({\bf $X$},{\bf $\hat X$}) = mean(({\bf $X$} - {\bf $\hat X$})$^2$)} \]
where the mean is here computed over all axes.

We observed the variance loss issue over a large variety of neural
architectures and datasets. 
In particular cases, we can also give a theoretical explanation
of the phenomenon, that looks strictly related to averaging. 
This is for instance the case of Principal Component Analysis (PCA), where the variance loss is 
precisely equal to the reconstruction error (it is well known that a {\em shallow}
Autoencoder implements PCA, see e.g \cite{DLbook}). 

Let us discuss this simple case first, since it helps to clarify
the issue.

Principal component analysis (PCA) is a well know statistical procedure 
for dimensionality reduction. The idea is to project data in a lower
dimensional space via an orthogonal linear transformation, choosing
the system of coordinates that maximize the variance of data (principal components). 
These are easily computed as the vectors with the largest eigenvalues
relative to the covariance matrix of the given dataset (centered around its mean points).
 \begin{figure}[ht]
     \centering
     \begin{minipage}{5.8cm}
     \includegraphics[width=.9\textwidth]{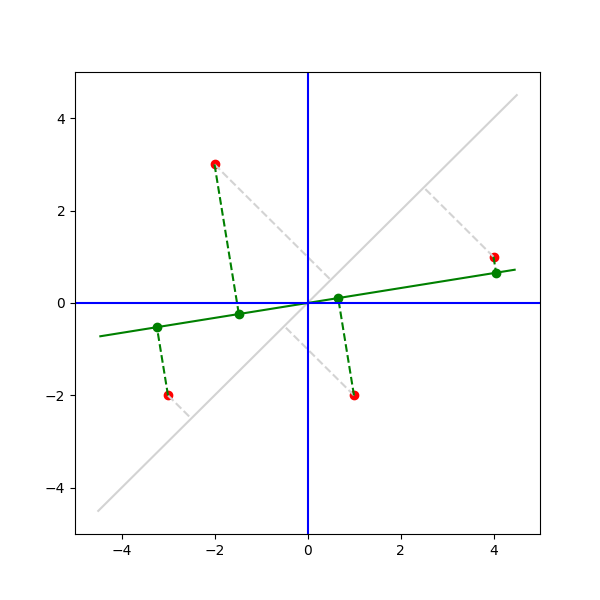}
     \captionof{figure}{
     The principal component is the green line. Projecting the red points
     on it, we maximize their variance or equivalently we minimize their
     quadratic distance.}
     \label{fig:pca}
     \end{minipage}\hspace{.4cm}
     \begin{minipage}{5.8cm}
     \includegraphics[width=.9\textwidth]{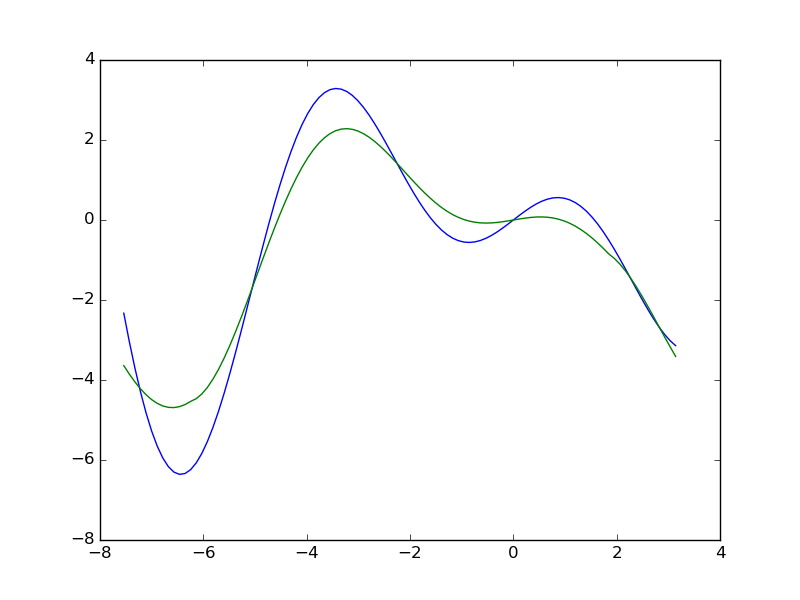}
     \caption{
     \label{fig:smoothing}The green line is a smoother version of the
     blue line, obtained by averaging values in a suitable neighborhood
     of each point. The two lines have the same mean; the mean squared
     error between them is 0.546,the variance loss is 2.648.}
     \end{minipage}
 \end{figure}
Since the distance of each point from the origin is fixed, by the
Pythagorean theorem, maximizing its variance is equivalent to minimize 
its quadratic error from the hyper-plane defined by the principal components. For the same reason, 
{\em the quadratic error of the reconstruction is equal to the sum of the
variance errors of the components which have been neglected}. 

This is a typical example of variance loss due to averaging. Since we 
want to renounce some components, the best we can do along them is to take the mean value. We entirely lose the variance along these directions, 
that is going to be paid in terms of reconstruction error.

\subsection{General case}
We expect to have a similar phenomenon even with more expressive 
networks. The idea is expressed in Figure~\ref{fig:smoothing}. Think
of the blue line as the real data manifold; due to averaging, the
network reconstructs a smoother version of the input data, resulting in a significant loss in terms of variance.

The need for averaging may have several motivations: it could be 
caused by a dimensionality reduction, as in the case of PCA, but also,
in the case of variational autoencoders, it could derive from the 
Gaussian sampling performed before reconstruction. Since the noise 
injected during sampling is completely unpredictable, the best the network can due is to reconstruct an ``average image'' corresponding to a
a portion of the latent space around the mean value $\mu_z(X)$, spanning an
area proportional to the variance $\sigma_z(X)^2$.

\begin{figure}[h!]
\hspace{.4cm}\includegraphics[width=.9\textwidth]{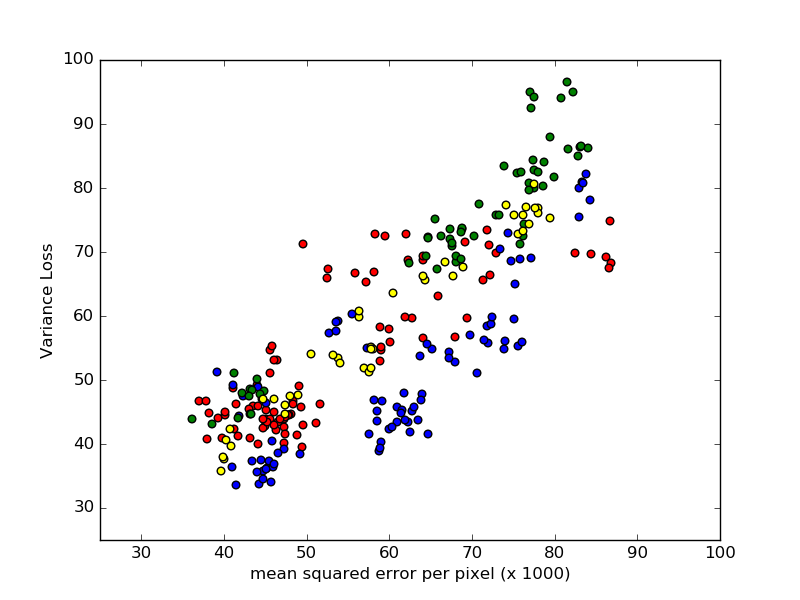}
\caption{\label{fig:var loss}Relation between mean squared error and variance loss. The different colors refer to different neural architectures: {\color{blue} $\bullet$} (blue) Dense Networks;  {\color{red}$\bullet$} (red) ResNet-like; {\color{green} $\bullet$} (green) Convolutional Networks; {\color {yellow} $\bullet$} Iterative Networks (DRAW-GQN-like)
}
\end{figure}

In Figure~\ref{fig:var loss}, we plot the relation between mean squared
error (mse) and variance loss for {\em reconstructed images}, computed over a large variety of different neural architectures and datasets: the distribution is close 
to the diagonal. Typically, the variance loss for {\em generated images} is
even greater. We must also account for a few pathological cases {\em not 
reported in the figure}, occurring with dense networks with very high 
capacity, and easily prone to overfitting. In this cases, mse is usually
relatively high, while variance loss may drop to 0.

In the general, deep case, however, it is not easy to relate the variance loss to the mean squared error. We just discuss a few cases.

If for each data $X$, the reconstructed value
$\hat{X}$ is comprised between $X$ and its mean value $\mu$, it is
easy to prove that the mean squared error is a {\em lower bound} to
the variance loss (the worse case is when $\hat{X} = \mu$, where the
variance loss is just equal to the mean squared error, as in the PCA
case). 

Similarly, let ${X_p}$ be an arbitrary permutation of elements of $X$
and let $\hat{X}= (X+X_p)/2$. Then, the mean square distance between 
$X$ and $\hat{X}$ is equal to the variance loss. However, the previous
property does not generalize when we average over an arbitrary number
of permutations; usually the mean squared error is lower than 
the quadratic distance between $X$ and $\hat{X}$, but we can also
get examples of the contrary.

We are still looking for a comfortable theoretical formulation of 
the property we are interested in.

\section{Addressing the variance loss}\label{sec:renormalization}
As we explained in the introduction, the variance loss issue has a great
practical relevance. Generative models are traditionally evaluated with metrics 
such as the popular Fr\'echet Inception Distance (FID)  aimed to compare
the distributions of real versus generated images trough a comparison of extracted
features. In the case of FID, the considered features are inception 
features; 
inception is usually preferred over other models due to the limited amount 
of preprocessing performed on input images. As a consequence, a bias
in generated data may easily result in a severe penalty in terms of FID score
(see \cite{FIDravaglia} for an extensive analysis of FID in relation to the training set).

The variance loss is particularly dangerous in a two stage setting \cite{TwoStage}, 
where a 
second VAE is used to sample in the latent space of the first VAE, in order to 
fix the possible mismatch between the aggregate inference distribution $Q(z)$ 
and the expected prior $P(z)$. 
The reduced variance induces a mismatch between the actual distribution of 
latent variables and those generated by the second VAE, hindering the beneficial 
effects of the second stage.

A simple way to address the variance loss issue consists in renormalizing generated 
data to match
the actual variance of real data by applying a multiplicative scaling factor. 
We implemented this simple approach in a variant of ours of the two stage model of Dai and Wipf, 
based on a new balancing strategy between reconstruction loss and
Kullback-Leibler described in \cite{balancing}. We refer to this latter work for 
details about the structure of the network, hyperparameter configuration, 
and training settings, clearly outside the scope of this article. The code is available at \href{https://github.com/asperti/BalancingVAE}{\url{https://github.com/asperti/BalancingVAE}}. 
\begin{figure}[h!]
\includegraphics[width=.95\columnwidth]{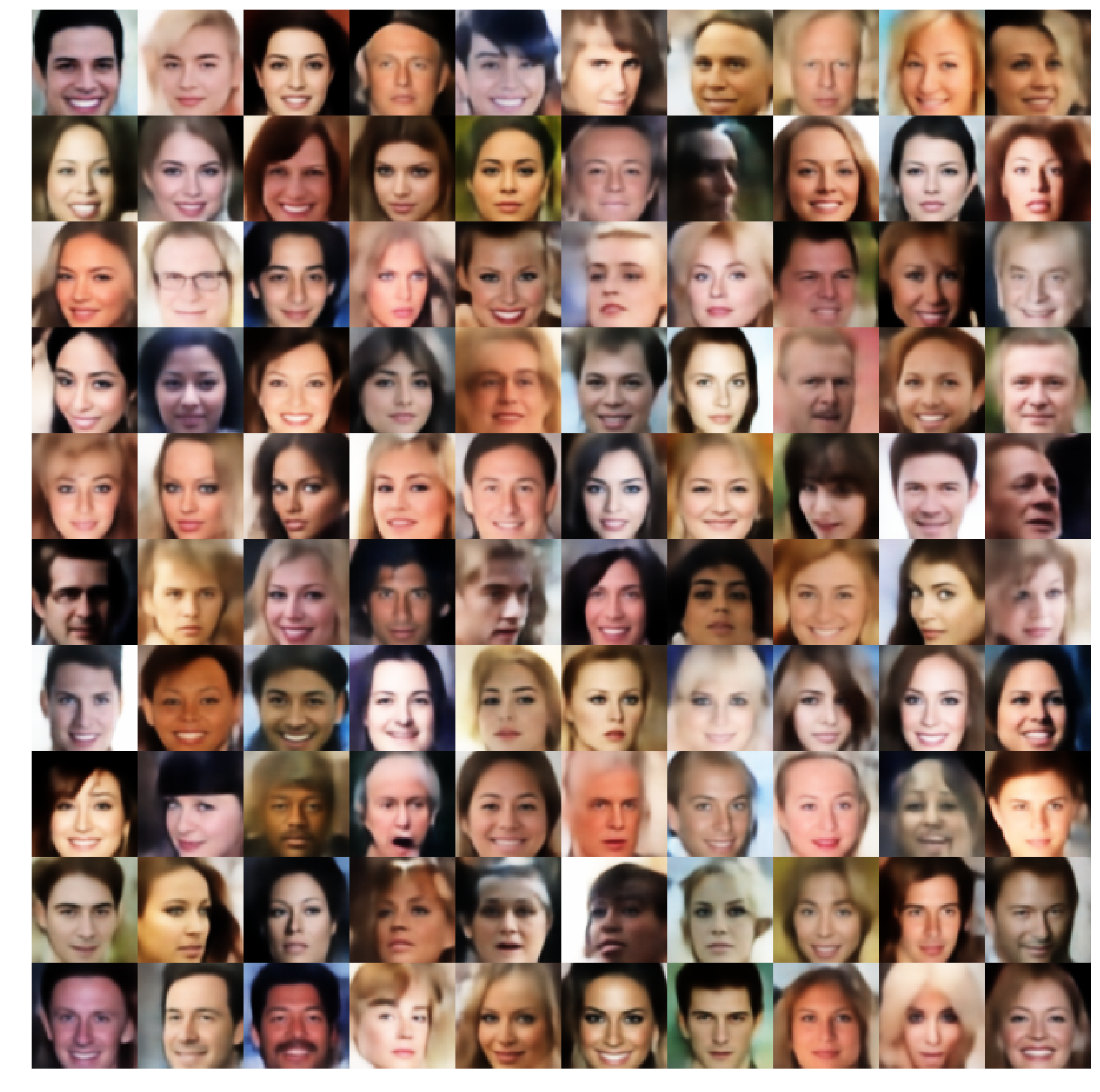}
\caption{\label{fig:celeba-gen}Examples of {\bf generated} faces. The resulting images do not show the
blurred appearance so typical of variational approaches, significantly improving their perceptive quality.}
\end{figure}
In Figure~\ref{fig:celeba-gen} we provide examples of randomly generated faces. Note
the particularly sharp quality of the images, so unusual for variational approaches.

Both for CIFAR-10 and CelebA, the renormalization operation results in an improvement in terms of FID scores, particularly significant in the case of CelebA, as reported in Tables~\ref{tab:results} and \ref{tab:CelebAresults}. At the best
of our knowledge, these are the best generative results ever obtained for these datasets without relying on adversarial training.
\begin{table}[h!]
\caption{\label{tab:results}CIFAR-10: summary of results}
\begin{center}
\begin{tabular}{|c|c|c|c|c|}\hline
      model   & REC & GEN-1 & GEN-2\\\hline
RAE-l2 \cite{deterministic} (128 vars)& $32.24 \pm ? $ & $80.8 \pm ? $ & $74.2 \pm ? $ \\\hline
2S-VAE \cite{TwoStage} & & $76.7 \pm 0.8$ & $72.9 \pm 0.9$\\\hline
2S-VAE (ours) & $53.8 \pm 0.9$ & $80.2 \pm 1.3$ & $ 69.8 \pm 1.1$\\
with normalization & $53.5 \pm 0.9$ & $78.6 \pm 1.2$ & ${\bf 69.4} \pm 1.0$\\\hline
\end{tabular}
\end{center}
\end{table}
\begin{table}[ht]
\caption{\label{tab:CelebAresults}CelebA: summary of results}
\begin{center}
\begin{tabular}{|c|c|c|c|c|}\hline
model        & REC & GEN-1 & GEN-2\\\hline
RAE-SN \cite{deterministic} & $36.0 \pm ? $ & $44.7 \pm ? $ & $40.9 \pm ? $ \\\hline
2S-VAE \cite{TwoStage} &     & $60.5 \pm 0.6$ & $44.4 \pm 0.7$ \\\hline
2S-VAE (ours) & $33.9 \pm 0.8$ & $43.6 \pm 1.3$ & $42.7 \pm 1.0$\\
with normalization & $33.7 \pm 0.8$ & $42.7 \pm 1.2$ & ${\bf 38.6} \pm 1.0$ \\\hline
\end{tabular}
\end{center}
\end{table}
In the Tables, we compare our generative model with the original two-stage model in \cite{TwoStage} and
with the recent deterministic model in \cite{deterministic}; as we mentioned above, 
these approaches represent the state of the art for generative models not based on adversarial 
training. For our model, we provide scores with and without normalization. For each model, we give FID scores for reconstructed images (REC), images generated after
the first stage (GEN-1), and images generated after the second stage (GEN-2). In the
case of the deterministic model \cite{deterministic}, the ``first stage" refers to sampling
after fitting a Gaussian on the latent space, where the second stage refers to a more
complex ex-post estimation of the latent space distribution via a GMM of ten Gaussians. The variance was computed over ten different 
trainings.

\begin{figure}[h!]
\includegraphics[width=.24\columnwidth]{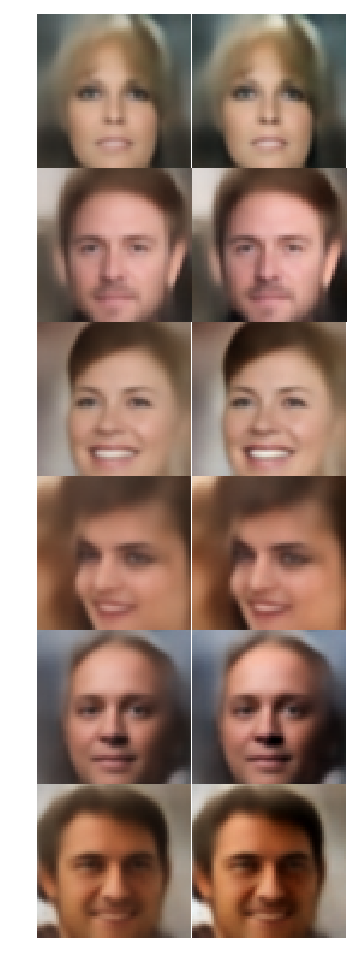}\includegraphics[width=.24\columnwidth]{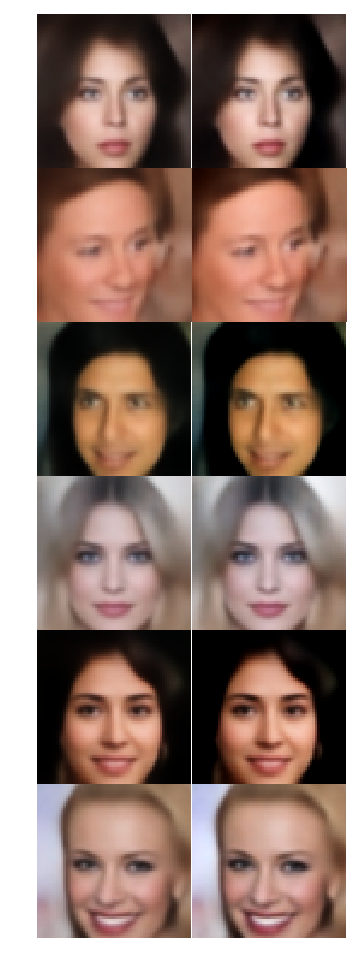}\includegraphics[width=.24\columnwidth]{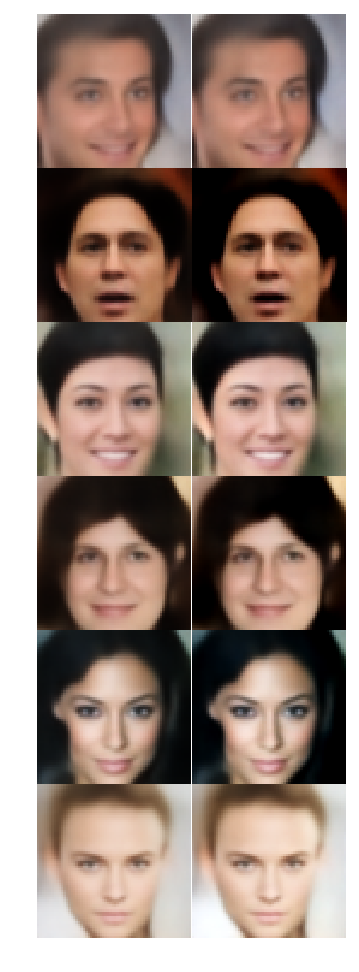}\includegraphics[width=.24\columnwidth]{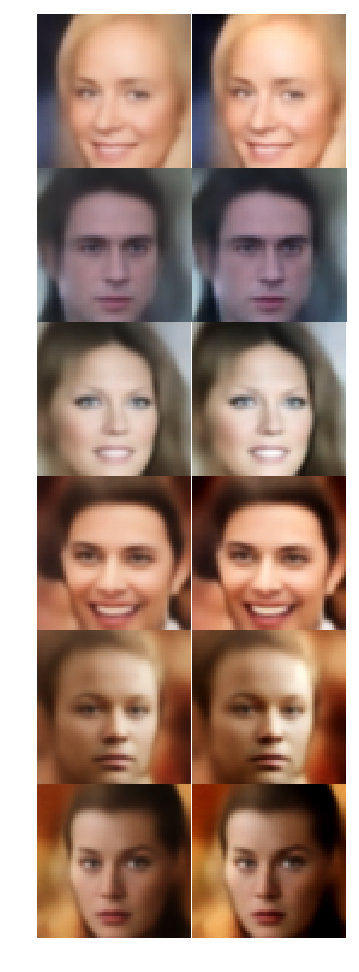}
\caption{\label{fig:varloss-gen}Faces with and without {\em latent space} re-normalization (right and left respectively). Images on the right have better contrasts and more definite contours.}
\end{figure}
In Figure~\ref{fig:varloss-gen} we show the difference
between faces generated from a same random seed with and
without latent space re-normalization. We hope that the quality of images
allows the reader to appreciate the improvement: renormalized
images (on the right) have more precise contours, 
sharper contrasts and more definite details.

\section{Conclusions}\label{Sec:conclusions}
In this article, we stressed an interesting and important 
problem typical of autoencoders and especially of variational
ones: the variance of generated data can be significantly lower than 
that of training data. We addressed the issue with a simple 
renormalization of generated data towards the expected moments
of the data distribution, permitting us to obtain significant
improvements in the quality of generated data, both in terms
of perceptual assessment and FID score. On typical datasets
such as CIFAR-10 and CelebA, this technique -
in conjunction with a new balancing strategy between 
reconstruction error and Kullback-Leibler divergence -
allowed us to get what seems to be the best generative results
ever obtained without the use of adversarial training.

\bibliographystyle{plain}
\bibliography{machine.bib,variational.bib}

\end{document}